\documentclass[final]{cvpr}

\usepackage{times}
\usepackage{epsfig}
\usepackage{graphicx}
\usepackage{amsmath}
\usepackage{amssymb}

\usepackage[utf8]{inputenc} %
\usepackage[T1]{fontenc}    %
\usepackage{url}            %
\usepackage{booktabs}       %
\usepackage{amsfonts}       %
\usepackage{nicefrac}       %
\usepackage{microtype}      %

\usepackage{subcaption}
\usepackage{cite}

\usepackage{floatrow}

\usepackage{graphicx}
\usepackage{amsmath}
\usepackage{amssymb}
\usepackage{multirow}
\usepackage{color}
\usepackage{wrapfig}

\makeatletter
\usepackage{xspace}
\def\@onedot{\ifx\@let@token.\else.\null\fi\xspace}
\DeclareRobustCommand\onedot{\futurelet\@let@token\@onedot}

\def\eg{\emph{e.g}\onedot} 
\def\ie{\emph{i.e}\onedot} 
 
\def\etc{\emph{etc}\onedot}

\usepackage[pagebackref=true,breaklinks=true,colorlinks,bookmarks=false]{hyperref}

\def\cvprPaperID{xxxx} 
\def\confYear{CVPR 2021}

\begin{document}

\title{DetectoRS: Detecting Objects with Recursive Feature Pyramid\\and Switchable Atrous Convolution}

\author{Siyuan Qiao\textsuperscript{1}~~~~~~Liang-Chieh Chen\textsuperscript{2}~~~~~~Alan Yuille\textsuperscript{1}\\
\textsuperscript{1}Johns Hopkins University\\\textsuperscript{2}Google Research\\
}

\maketitle

\begin{abstract}
Many modern object detectors demonstrate outstanding performances by using the mechanism of looking and thinking twice.
In this paper, we explore this mechanism in the backbone design for object detection.
At the macro level, we propose Recursive Feature Pyramid, which incorporates extra feedback connections from Feature Pyramid Networks into the bottom-up backbone layers.
At the micro level, we propose Switchable Atrous Convolution, which convolves the features with different atrous rates and gathers the results using switch functions.
Combining them results in DetectoRS, which significantly improves the performances of object detection.
On COCO test-dev, DetectoRS achieves state-of-the-art 55.7\% box AP for object detection, 48.5\% mask AP for instance segmentation, and 50.0\% PQ for panoptic segmentation.
The code is made publicly available\footnote{ \url{https://github.com/joe-siyuan-qiao/DetectoRS}}.
\end{abstract}
\section{Introduction}

\begin{figure*}[t]
    \centering
    \begin{subfigure}[b]{0.44\textwidth}
         \centering
         \includegraphics[width=\textwidth]{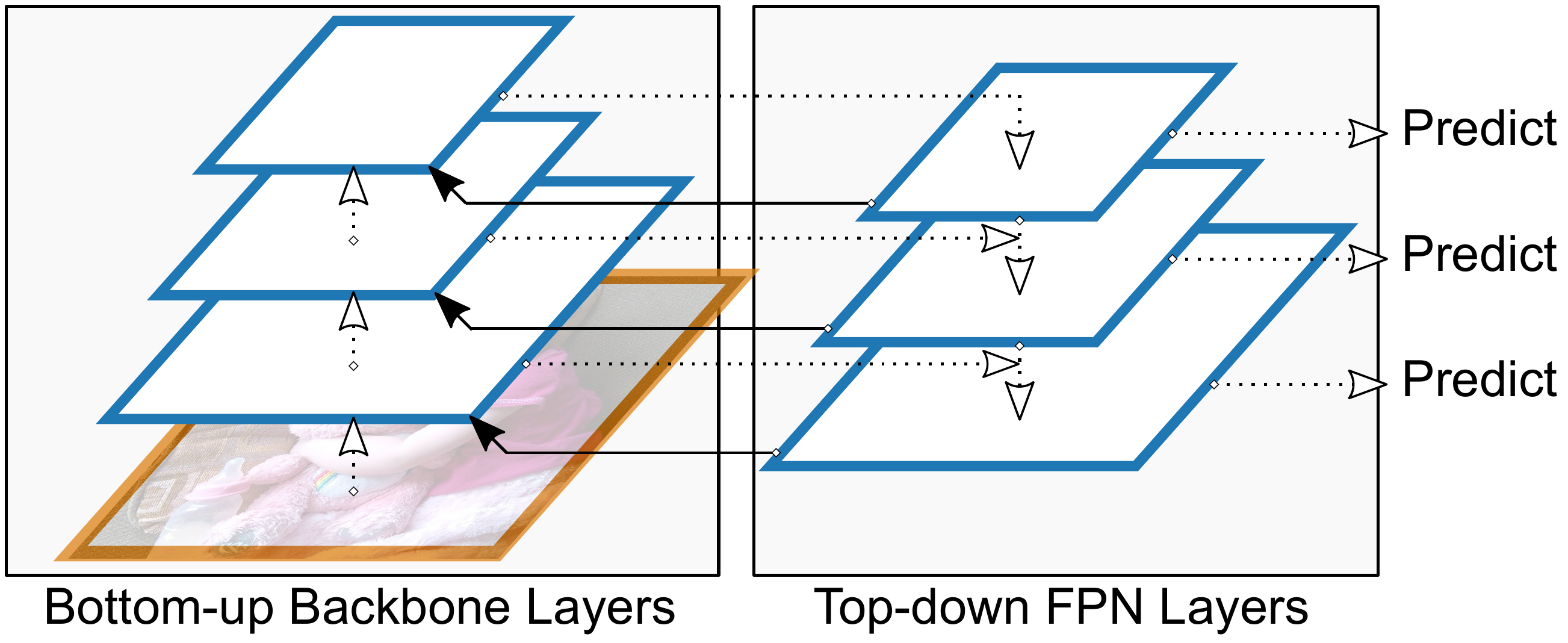}
         \caption{Macro Design: Recursive Feature Pyramid.}
         \label{fig:rfP_intro}
     \end{subfigure}
     \hfill
     \begin{subfigure}[b]{0.55\textwidth}
         \centering
         \includegraphics[width=\textwidth]{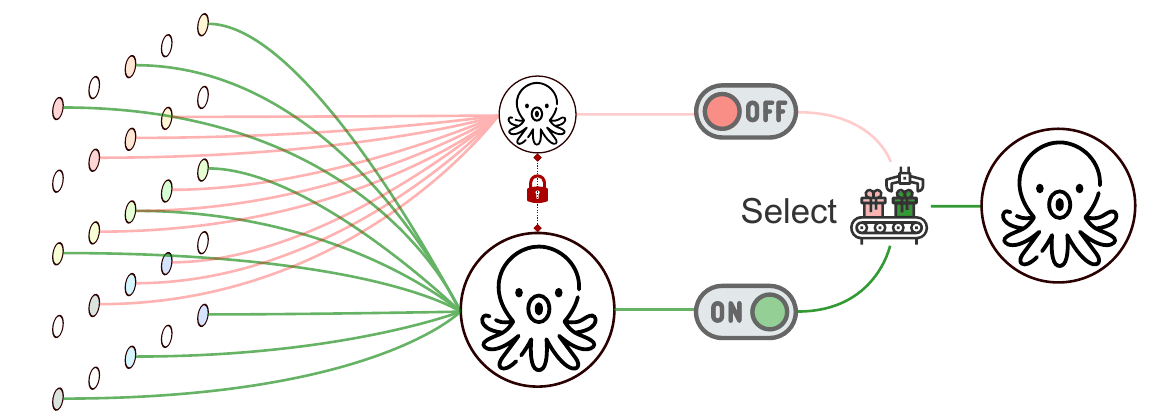}
         \caption{Micro Design: Switchable Atrous Convolution.}
         \label{fig:sac_intro}
     \end{subfigure}
     \caption{(a) Our Recursive Feature Pyramid adds feedback connections (solid lines) from the top-down FPN layers to the bottom-up backbone layers to look at the image twice or more.
     (b) Our Switchable Atrous Convolution looks twice at the input features with different atrous rates and the outputs are combined together by switches.
     }
\end{figure*}

To detect objects, human visual perception selectively enhances and suppresses neuron activation by passing high-level semantic information through feedback connections \cite{beck2009top,desimone1998visual,desimone1995neural}.
Inspired by the human vision system, the mechanism of \textit{looking and thinking twice} has been instantiated in computer vision, and demonstrated outstanding performance~\cite{cao2015look,fasterrcnn,cascadercnn}.
Many popular two-stage object detectors, \eg, Faster R-CNN~\cite{fasterrcnn}, output object proposals first, based on which regional features are then extracted to detect objects.
Following the same direction, Cascade R-CNN~\cite{cascadercnn} develops a multi-stage detector, where subsequent detector heads are trained with more selective examples. 
The success of this design philosophy motivates us to explore it in the neural network backbone design for object detection.
In particular, we deploy the mechanism at both the macro and micro levels, resulting in our proposed DetectoRS which significantly improves the performance of the state-of-art object detector HTC~\cite{htc} by a great margin while a similar inference speed is maintained, as shown in \tabref{tab:intro}.

At the macro level, our proposed Recursive Feature Pyramid (RFP) builds on top of the Feature Pyramid Networks (FPN)~\cite{fpn} by incorporating extra feedback connections from the FPN layers into the bottom-up backbone layers, as illustrated in \figref{fig:rfP_intro}. Unrolling the recursive structure to a sequential implementation, we obtain a backbone for object detector that looks at the images twice or more. Similar to the cascaded detector heads in Cascade R-CNN trained with more selective examples, our RFP recursively enhances FPN to generate increasingly powerful representations.
Resembling Deeply-Supervised Nets~\cite{lee2015deeply}, the feedback connections bring the features that directly receive gradients from the detector heads back to the low levels of the bottom-up backbone to speed up training and boost performance.
Our proposed RFP implements a sequential design of \textit{looking and thinking twice}, where the bottom-up backbone and FPN are run multiple times with their output features dependent on those in the previous steps.

\begin{table}[t]
\small
\setlength{\tabcolsep}{0.6em}
    \centering
    \begin{tabular}{ll|ccc}
    \toprule
    Method & Backbone & AP$_{\text{box}}$ & AP$_{\text{mask}}$ & FPS\\
    \midrule
    HTC~\cite{htc} & ResNet-50 & 43.6 & 38.5 & 4.3\\
    DetectoRS & ResNet-50 & 51.3 & 44.4 & 3.9\\
    \bottomrule
    \end{tabular}
    \caption{A glimpse of the improvements of the box and mask AP by our DetectoRS on COCO \texttt{test-dev}.}
    \label{tab:intro}
\end{table}

At the micro level, we propose Switchable Atrous Convolution (SAC), which convolves the same input feature with different atrous rates~\cite{holschneider1989real,papandreou2014untangling,chen2014semantic} and gathers the results using switch functions.
Fig.~\ref{fig:sac_intro} shows an illustration of the concept of SAC.
The switch functions are spatially dependent, \ie, each location of the feature map might have different switches to control the outputs of SAC.
To use SAC in the detector, we convert all the standard 3x3 convolutional layers in the bottom-up backbone to SAC, which improves the detector performance by a large margin.
Some previous methods adopt conditional convolution, \eg, \cite{condconv,li2019selective}, which also combines results of different convolutions as a single output.
Unlike those methods whose architecture requires to be trained from scratch, SAC provides a mechanism to easily convert pretrained standard convolutional networks (\eg, ImageNet-pretrained~\cite{imagenet} checkpoints).
Moreover, a new weight locking mechanism is used in SAC where the weights of different atrous convolutions are the same except for a trainable difference.

Combining the proposed RFP and SAC results in our DetectoRS. To demonstrate its effectiveness, we incorporate DetectoRS into the state-of-art HTC~\cite{htc} on the challenging COCO dataset~\cite{coco}.
On COCO \texttt{test-dev}, we report box AP for object detection~\cite{everingham2015pascal}, mask AP for instance segmentation~\cite{hariharan2014simultaneous}, and PQ for panoptic segmentation~\cite{kirillov2019panoptic}. DetectoRS with ResNet-50~\cite{resnet} as backbone significantly improves HTC~\cite{htc} by 7.7\% box AP and 5.9\% mask AP. Additionally, equipping our DetectoRS with ResNeXt-101-64x4d~\cite{resnext} achieves state-of-the-art 55.7\% box AP and 48.5\% mask AP.
Together with the stuff prediction from DeepLabv3+~\cite{deeplabv3plus2018} with Wide-ResNet-41~\cite{sslpanopticdeeplab2020} as backbone, DetectoRS sets a new record of 50.0\% PQ for panoptic segmentation.

\section{Related Works}

\noindent\textbf{Object Detection}.
There are two main categories of object detection methods: one-stage methods, \eg, \cite{sermanet2013overfeat,liu2016ssd,redmon2017yolo9000,lin2017focal,zhang2018single,zhao2019m2det,wang2020fcos,li2020learning},
and multi-stage methods, \eg, \cite{girshick2015fast,fasterrcnn,maskrcnn,chen2018masklab,cascadercnn,htc,cao2020d2det,wu2020rethinking,guo2020augfpn,jiang2020sp}.
Multi-stage detectors are usually more flexible and accurate but more complex than one-stage detectors.
In this paper, we use a multi-stage detector HTC~\cite{htc} as our baseline and show comparisons with both categories.

\noindent\textbf{Multi-Scale Features}.
Our Recursive Feature Pyramid is based on Feature Pyramid Networks (FPN)~\cite{fpn}, an effective object detection system that exploits multi-scale features.
Previously, many object detectors directly use the multi-scale features extracted from the backbone \cite{liu2016ssd,cai2016unified}, while FPN incorporates a top-down path to sequentially combine features at different scales. PANet~\cite{panet} adds another bottom-up path on top of FPN. STDL~\cite{zhou2018scale} proposes to exploit cross-scale features by a scale-transfer module.
G-FRNet~\cite{amirul2017gated} adds feedback with gating units.
NAS-FPN~\cite{nasfpn} and Auto-FPN~\cite{xu2019auto} use neural architecture search~\cite{zoph2016neural} to find the optimal FPN structure.
EfficientDet~\cite{efficientdet} proposes to repeat a simple BiFPN layer.
Unlike them, our proposed Recursive Feature Pyramid goes through the bottom-up backbone repeatedly to enrich the representation power of FPN.
Additionally, we incorporate the Atrous Spatial Pyramid Pooling (ASPP)~\cite{chen2017deeplabv3,deeplabv3plus2018} into FPN to enrich features, similar to the mini-DeepLab design in Seamless~\cite{porzi2019seamless}.

\noindent\textbf{Recursive Convolutional Network}.
Many recursive methods have been proposed to address different types of computer vision problems, \eg, \cite{liang2015recurrent,kim2016deeply,tai2017image}.
Recently, a recursive method CBNet~\cite{liu2019cbnet} is proposed for object detection, which cascades multiple backbones to output features as the input of FPN.
By contrast, our RFP performs recursive computations with proposed ASPP-enriched FPN {\it included} along with effective fusion modules.

\noindent\textbf{Conditional Convolution}
Conditional convolutional networks adopt dynamic kernels, widths, or depths, \eg, \cite{lin2017runtime,liu2018dynamic,yu2018slimmable,li2019selective,chen2019dynamic,condconv}.
Unlike them, our proposed Switchable Atrous Convolution (SAC) allows an effective conversion mechanism from standard convolutions to conditional convolutions without changing any pretrained models.
SAC is thus a plug-and-play module for many pretrained backbones.
Moreover, SAC uses global context information and a novel weight locking mechanism to make it more effective.

\section{Recursive Feature Pyramid}\label{sec:rfp}

\begin{figure*}
    \centering
    \begin{subfigure}[t]{0.45\textwidth}
         \centering
         \includegraphics[width=\textwidth]{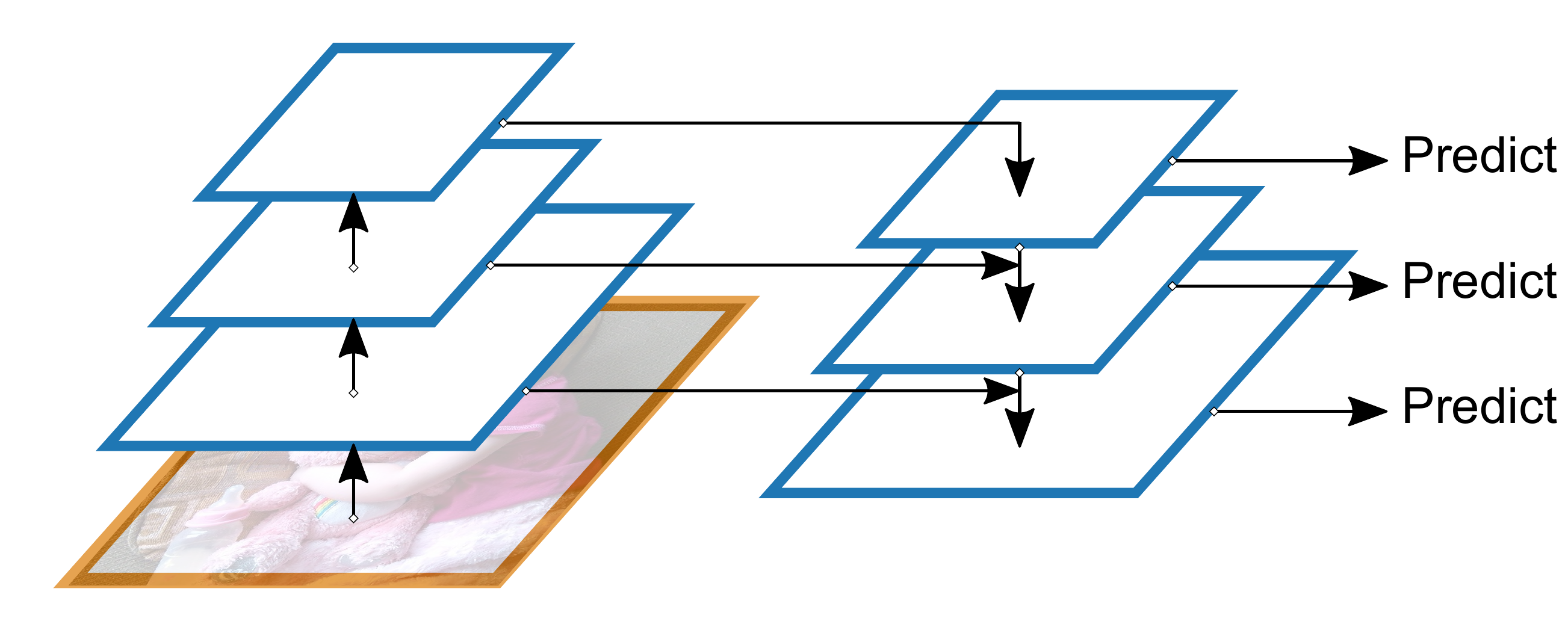}
         \caption{FPN.}
         \label{fig:fpn}
     \end{subfigure}
     \hfill
     \begin{subfigure}[t]{0.45\textwidth}
         \centering
         \includegraphics[width=\textwidth]{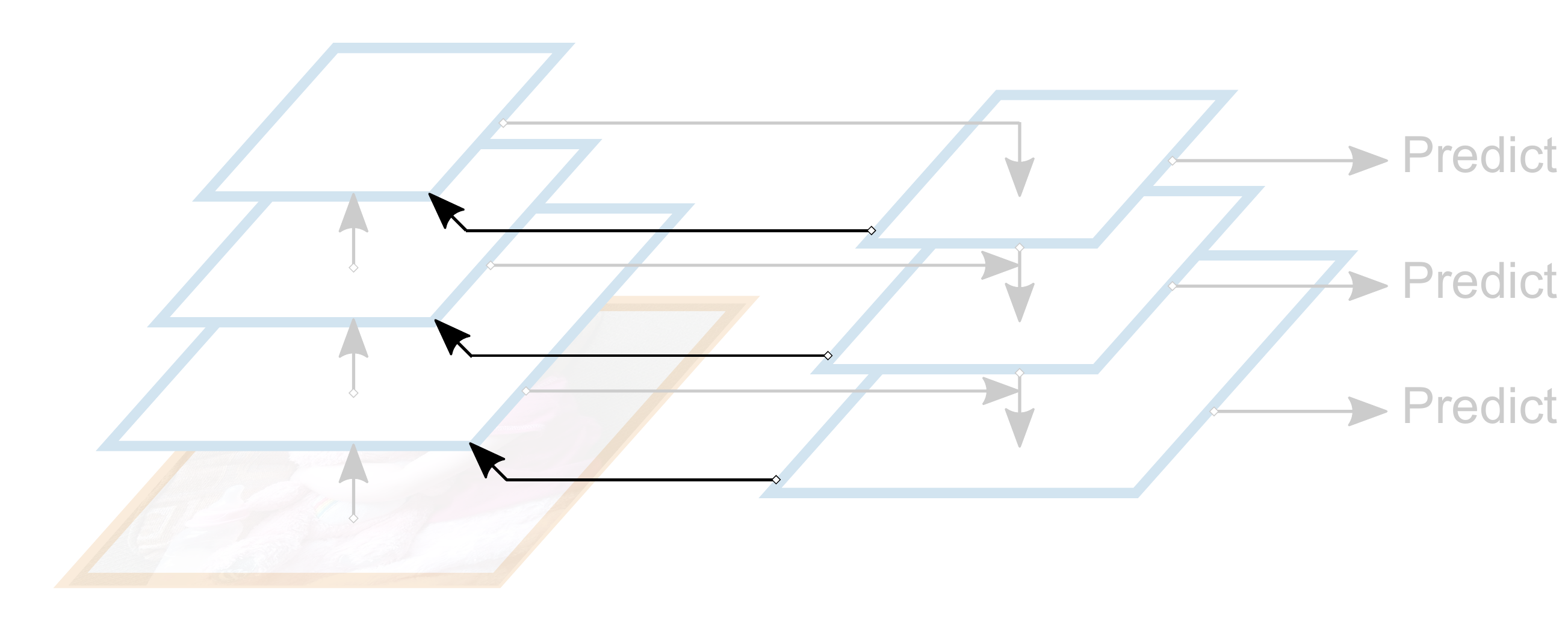}
         \caption{RFP adds feedback connections to FPN.}
         \label{fig:rfp}
     \end{subfigure}\\
     \begin{subfigure}[t]{0.8\textwidth}
         \centering
         \includegraphics[width=\textwidth]{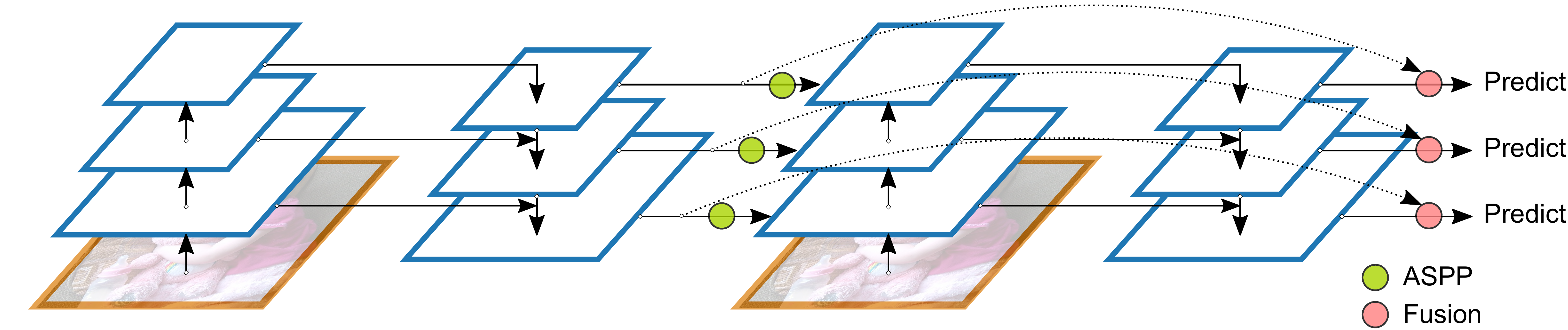}
         \caption{Unrolling RFP to an example 2-step sequential implementation.}
         \label{fig:rfp_unrolled}
     \end{subfigure}
     \caption{The architecture of Recursive Feature Pyramid (RFP). (a) Feature Pyramid Networks (FPN). (b) Our RFP incorporates feedback connections into FPN. (c) RFP unrolled to a 2-step sequential network.
     }
     \label{fig:rfp_overview}
\end{figure*}

\subsection{Feature Pyramid Networks}
This subsection provides the background of Feature Pyramid Networks (FPN).
Let $\mathbf{B}_i$ denote the $i$-th stage of the bottom-up backbone, and $\mathbf{F}_i$ denote the $i$-th top-down FPN operation. The backbone equipped with FPN outputs a set of feature maps \{$\mathbf{f}_i$~|~$i=1,...,S$\}, where $S$ is the number of the stages.
For example, $S=3$ in ~\figref{fig:fpn}.
$\forall i=1,...,S$, the output feature $\mathbf{f}_i$ is defined by
\begin{equation}
    \mathbf{f}_i = \mathbf{F}_i\big(\mathbf{f}_{i+1}, \mathbf{x}_i\big),~~~\mathbf{x}_i = \mathbf{B}_i\big(\mathbf{x}_{i-1}\big),
\end{equation}
where $\mathbf{x}_0$ is the input image and $\mathbf{f}_{S+1}=\mathbf{0}$.
The object detector built on FPN uses $\mathbf{f}_i$ for the detection computations.

\subsection{Recursive Feature Pyramid}

Our proposed Recursive Feature Pyramid (RFP) adds feedback connections to FPN as highlighted in ~\figref{fig:rfp}.
Let $\mathbf{R}_i$ denote the feature transformations before connecting them back to the bottom-up backbone.
Then, $\forall i=1,...,S$, the output feature $\mathbf{f_i}$ of RFP is defined by
\begin{equation}
    \mathbf{f}_i = \mathbf{F}_i\big(\mathbf{f}_{i+1}, \mathbf{x}_i\big),~~~\mathbf{x}_i = \mathbf{B}_i\big(\mathbf{x}_{i-1}, \mathbf{R}_i(\mathbf{f}_i)\big),
\end{equation}
which makes RFP a recursive operation.
We unroll it to a sequential network, \ie,
$\forall i=1,...,S, t=1,...T$,
\begin{equation}\label{eq:rfp}
    \mathbf{f}^t_i = \mathbf{F}^t_i\big(\mathbf{f}^t_{i+1}, \mathbf{x}^t_i\big),~\mathbf{x}^t_i = \mathbf{B}^t_i\big(\mathbf{x}^t_{i-1}, \mathbf{R}^t_i(\mathbf{f}_i^{t-1})\big),
\end{equation}
where $T$ is the number of unrolled iterations, and we use superscript $t$ to denote operations and features at the unrolled step $t$.
$\mathbf{f}_i^0$ is set to $\mathbf{0}$. 
In our implementation, $\mathbf{F}_i^t$ and $\mathbf{R}_i^t$ are shared across different steps.
We show both shared and different $\mathbf{B}_i^t$ in the ablation study in ~\secref{sec:exp} as well as the performances with different $T$'s.
In our experiments, we use different $\mathbf{B}_i^t$ and set $T=2$, unless otherwise stated.

We make changes to the ResNet~\cite{resnet} backbone $\mathbf{B}$ to allow it to take both $\mathbf{x}$ and $\mathbf{R}(\mathbf{f})$ as its input.
ResNet has four stages, each of which is composed of several similar blocks.
We only make changes to the first block of each stage, as shown in \figref{fig:resnet}.
This block computes a 3-layer feature and adds it to a feature computed by a shortcut.
To use the feature $\mathbf{R}(\mathbf{f})$, we add another convolutional layer with the kernel size set to $1$.
The weight of this layer is initialized with $\mathbf{0}$ to make sure it does not have any real effect when we load the weights from a pretrained checkpoint.

\subsection{ASPP as the Connecting Module}

\begin{figure}[!htp]
    \centering
    \includegraphics[width=\linewidth]{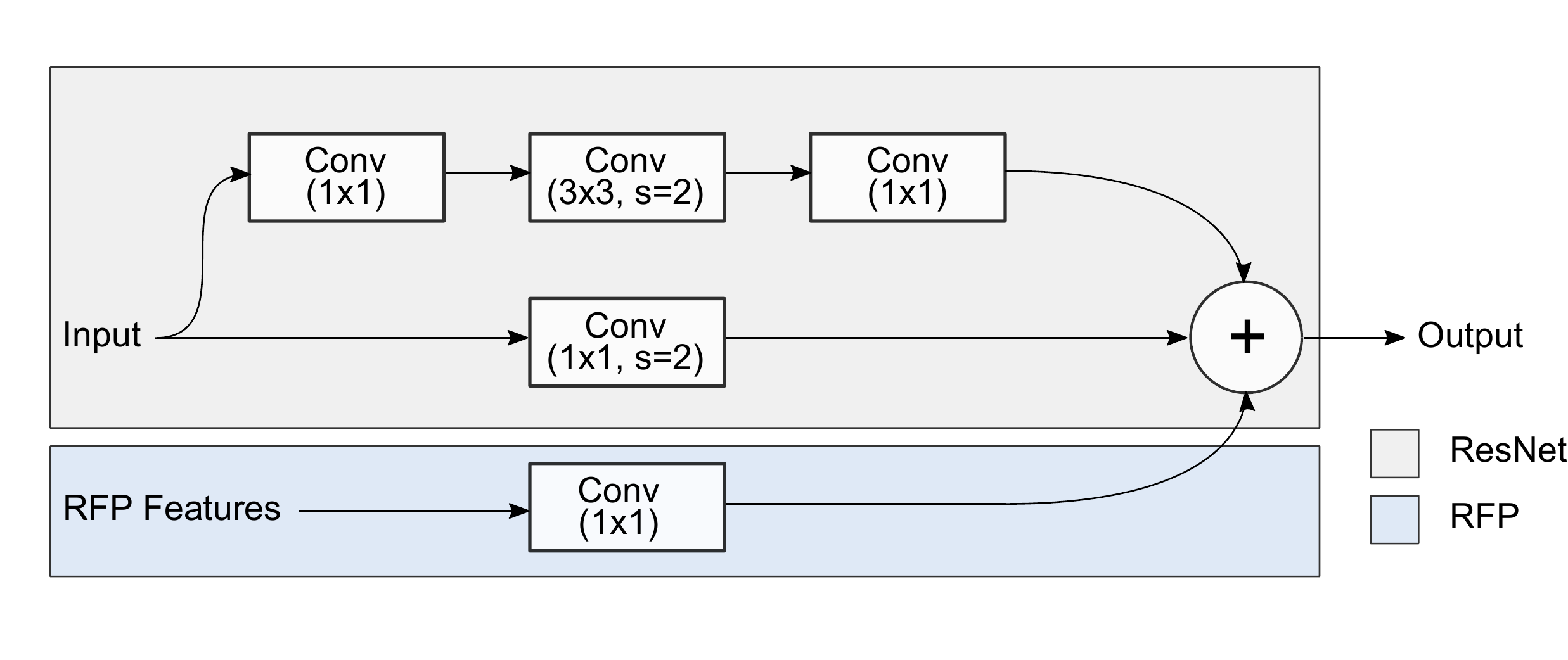}
    \caption{RFP adds transformed features to the first block of each stage of ResNet.}
    \label{fig:resnet}
\end{figure}

\begin{figure*}
    \centering
    \includegraphics[width=\linewidth]{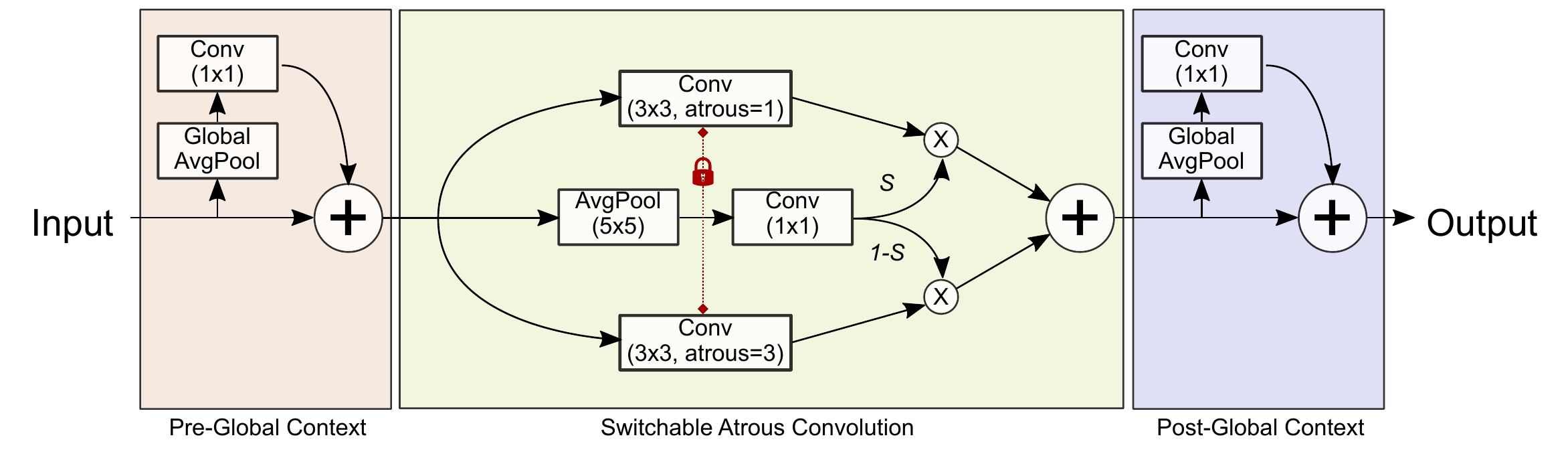}
    \caption{Switchable Atrous Convolution (SAC).
    We convert every 3x3 convolutional layer in the backbone ResNet to SAC, which softly switches the convolutional computation between different atrous rates.
    The \textbf{lock} indicates that the weights are the same except for a trainable difference (see Eq.~\ref{eq:sac}).
    Two global context modules add image-level information to the features. 
    }
    \label{fig:sac}
\end{figure*}

We use Atrous Spatial Pyramid Pooling (ASPP)~\cite{chen2017deeplab} to implement the connecting module $\mathbf{R}$, which takes a feature $\mathbf{f}^t_i$ as its input and transforms it to the RFP feature used in ~\figref{fig:resnet}.
In this module, there are four parallel branches that take $\mathbf{f}^t_i$ as their inputs, the outputs of which are then concatenated together along the channel dimension to form the final output of $\mathbf{R}$.
Three branches of them use a convolutional layer followed by a ReLU layer, the number of the output channels is $1/4$ the number of the input channels.
The last branch uses a global average pooling layer to compress the feature, followed by a 1x1 convolutional layer and a ReLU layer to transform the compressed feature to a $1/4$-size (channel-wise) feature. Finally, it is resized and concatenated with the features from the other three branches.
The convolutional layers in those three branches are of the following configurations: kernel size = [1, 3, 3], atrous rate = [1, 3, 6], padding = [0, 3, 6].
Unlike the original ASPP~\cite{chen2017deeplab}, we do not have a convolutional layer following the concatenated features as in here $\mathbf{R}$ does not generate the final output used in dense prediction tasks.
Note that each of the four branches yields a feature with channels $1/4$ that of the input feature, and concatenating them generates a feature that has the same size as the input feature of $\mathbf{R}$.
In ~\secref{sec:exp}, we show the performances of RFP with and without ASPP module.

\subsection{Output Update by the Fusion Module}

\begin{figure}[t]
    \centering
    \includegraphics[width=\linewidth]{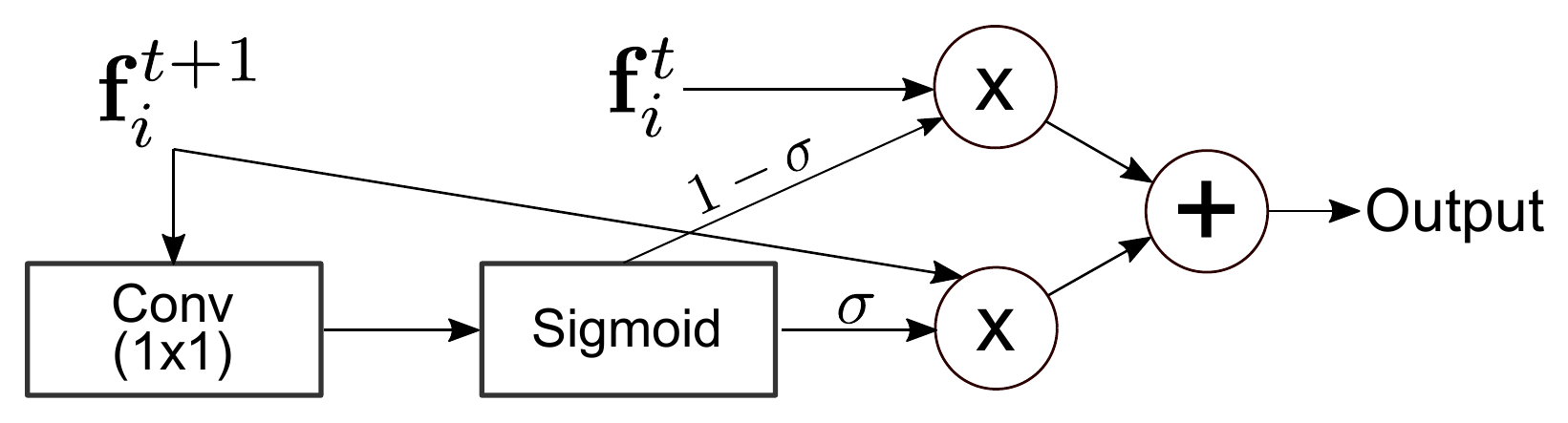}
    \caption{The fusion module used in RFP.
    $\sigma$ is the output of Sigmoid, which is used to fuse features from different steps.
    }
    \label{fig:weight}
\end{figure}

As shown in ~\figref{fig:rfp_unrolled}, our RFP additionally uses a fusion module to combine $\mathbf{f}_i^t$ and $\mathbf{f}_i^{t+1}$ to update the values of $\mathbf{f}_i$ at the unrolled stage $t+1$ used in ~\equref{eq:rfp}.
The fusion module is very similar to the update process in recurrent neural networks~\cite{lstm} if we consider $\mathbf{f}_i^t$ as a sequence of data.
The fusion module is used for unrolled steps from 2 to $T$.
At the unrolled step $t+1$ ($t=1,...,T-1$), the fusion module takes the feature $\mathbf{f}_i^t$ at the step $t$ and the feature $\mathbf{f}_i^{t+1}$ newly computed by FPN at the step $t+1$ as its input.
The fusion module uses the feature $\mathbf{f}_i^{t+1}$ to compute an attention map by a convolutional layer followed by a Sigmoid operation.
The resulting attention map is used to compute the weighted sum of $\mathbf{f}_i^{t}$ and $\mathbf{f}_i^{t+1}$ to form an updated $\mathbf{f}_i$.
This $\mathbf{f}_i$ will be used as $\mathbf{f}_i^{t+1}$ for the computation in the following steps.
In the ablation study in ~\secref{sec:exp}, we will show the performances of RFP with and without the fusion module.

\begin{table*}
\small
\setlength{\tabcolsep}{0.33em}
    \centering
    \begin{tabular}{ccc|cccccc|cccccc|c}
    \toprule
    & & & \multicolumn{6}{c|}{Box} & \multicolumn{6}{c|}{Mask} & Runtime \\
    HTC & RFP & SAC & AP & AP$_{50}$ & AP$_{75}$ & AP$_{S}$ & AP$_{M}$ & AP$_{L}$ & AP & AP$_{50}$ & AP$_{75}$ & AP$_{S}$ & AP$_{M}$ & AP$_{L}$ & FPS\\
    \midrule
    \checkmark & &  &  42.0 & 60.8 & 45.5 & 23.7 & 45.5 & 56.4 & 37.1 & 58.2 & 39.9 & 19.1 & 40.2 & 51.9 & 4.3\\
    \checkmark & \checkmark & & 46.2 & 65.1 & 50.2 & 27.9 & 50.3 & 60.3 & 40.4 & 62.5 & 43.5 & 22.3 & 43.8 & 54.9 & 4.1 \\
    \checkmark & & \checkmark & 46.3 & 65.8 & 50.2 & 27.8 & 50.6 & 62.4 & 40.4 & 63.1 & 43.4 & 22.7 & 44.2 & 56.4 & 4.2 \\
    \checkmark & \checkmark & \checkmark & 49.0 & 67.7 & 53.0 & 30.1 & 52.6 & 64.9 & 42.1 & 64.8 & 45.5 & 23.9 & 45.6 & 57.8 & 3.9 \\
    \bottomrule
    \end{tabular}
    \caption{Detection results on COCO \texttt{val2017} with ResNet-50 as backbone.
    The models are trained for 12 epochs.}
    \label{tab:rfp_sac}
\end{table*}

\section{Switchable Atrous Convolution}\label{sec:sac}

\subsection{Atrous Convolution}

Atrous convolution~\cite{holschneider1989real,papandreou2014untangling,chen2014semantic} is an effective technique to enlarge
the field-of-view of filters at any convolutional layer. In particular, atrous convolution with atrous rate $r$ introduces $r-1$ zeros between consecutive filter values, equivalently enlarging the kernel size of a $k\times k$ filter
to $k_e=k+(k-1)(r-1)$ without increasing the number of parameters or the amount of computation.
Fig.~\ref{fig:sac_intro} shows an example of a 3x3 convolutional layer with the atrous rate set to 1 (red) and 2 (green):
the same kind of object of different scales could be roughly detected by the same set of convolutional weights using different atrous rates.

\subsection{Switchable Atrous Convolution}

In this subsection, we present the details of our proposed Switchable Atrous Convolution (SAC).
Fig.~\ref{fig:sac} shows the overall architecture of SAC, which has three major components: two global context modules appended \textit{before} and \textit{after} the SAC component.
This subsection focuses on the main SAC component in the middle and we will explain the global context modules afterwards.

We use $\mathbf{y}=\text{\bf Conv}(\mathbf{x}, \mathbf{w}, r)$ to denote the convolutional operation with weight $\mathbf{w}$ and atrous rate $r$ which takes $\mathbf{x}$ as its input and outputs $\mathbf{y}$.
Then, we can convert a convolutional layer to SAC as follows.
\begin{align}\label{eq:sac}
\begin{split}
    \text{\bf Conv}\big(\mathbf{x}, \mathbf{w}, 1\big)\xrightarrow[\text{to SAC}]{\text{Convert}}
    \mathbf{S}\big(\mathbf{x}\big)\cdot\text{\bf Conv}\big(\mathbf{x}, \mathbf{w}, 1\big)\\ + \big(1-\mathbf{S}(\mathbf{x})\big)\cdot \text{\bf Conv}\big(\mathbf{x}, \mathbf{w} + \Delta\mathbf{w}, r\big)
\end{split}
\end{align}
where $r$ here is a hyper-parameter of SAC, $\Delta\mathbf{w}$ is a trainable weight, and the switch function $\mathbf{S}(\cdot)$ is implemented as an average pooling layer with a 5x5 kernel followed by a 1x1 convolutional layer (see Fig.~\ref{fig:sac}).
The switch function is input and location dependent; thus, the backbone model is able to adapt to different scales as needed.
We set $r=3$ in our experiments, unless stated otherwise.

We propose a locking mechanism by setting one weight as $\mathbf{w}$ and the other as $\mathbf{w} + \Delta\mathbf{w}$ for the following reasons.
Object detectors usually use pretrained checkpoints to initialize the weights.
However, for an SAC layer converted from a standard convolutional layer, the weight for the larger atrous rate is missing.
Since objects at different scales can be roughly detected by the same weight with different atrous rates, it is natural to initialize the missing weights with those in the pretrained model.
Our implementation uses $\mathbf{w} + \Delta\mathbf{w}$ for the missing weight where $\mathbf{w}$ is from the pretrained checkpoint and  $\Delta\mathbf{w}$ is initialized with $\mathbf{0}$.
When fixing $\Delta\mathbf{w}=0$, we observe a drop of $0.1$\% AP.
But $\Delta\mathbf{w}$ alone without the locking mechanism degrades AP a lot.

\subsection{Global Context}
As shown in Fig.~\ref{fig:sac}, we insert two global context modules before and after the main component of SAC.
These two modules are light-weighted as the input features are first compressed by a global average pooling layer.
The global context modules are similar to SENet~\cite{senet} except for two major differences: (1) we only have one convolutional layer without any non-linearity layers, and (2) the output is added back to the main stream instead of multiplying the input by a re-calibrating value computed by Sigmoid.
Experimentally, we found that adding the global context information before the SAC component (\ie, adding global information to the switch function) has a positive effect on the detection performance. 
We speculate that this is because $\mathbf{S}$ can make more stable switching predictions when global information is available.
We then move the global information outside the switch function and place it before and after the major body so that both $\text{\bf Conv}$ and $\mathbf{S}$ can benefit from it.
We did not adopt the original SENet formulation as we found no improvement on the final model AP.
In the ablation study in Sec.~\ref{sec:exp}, we show the performances of SAC with and without the global context modules.

\subsection{Implementation Details}
In our implementation, we use deformable convolution~\cite{dcn,dcnv2} to replace both of the convolutional operations in Eq.~\ref{eq:sac}.
The offset functions of them are not shared, which are initialized to predict $\mathbf{0}$ when loading from a pretrained backbone.
Experiments in ~\secref{sec:exp} will show performance comparisons of SAC with and without deformable convolution.
We adopt SAC on ResNet and its variants \cite{resnet,resnext} by replacing all the 3x3 convolutional layers in the backbone.
The weights and the biases in the global context modules are initialized with $\mathbf{0}$.
The weight in the switch $\mathbf{S}$ is initialized with $\mathbf{0}$ and the bias is set to $\mathbf{1}$.
$\Delta\mathbf{w}$ is initialized with $\mathbf{0}$.
The above initialization strategy guarantees that when loading the backbone pretrained on ImageNet~\cite{imagenet}, converting all the 3x3 convolutional layers to SAC will not change the output before taking any steps of training on COCO~\cite{coco}.
\begin{figure*}
    \centering
    \begin{subfigure}[t]{\linewidth}
         \centering
         \includegraphics[width=\linewidth]{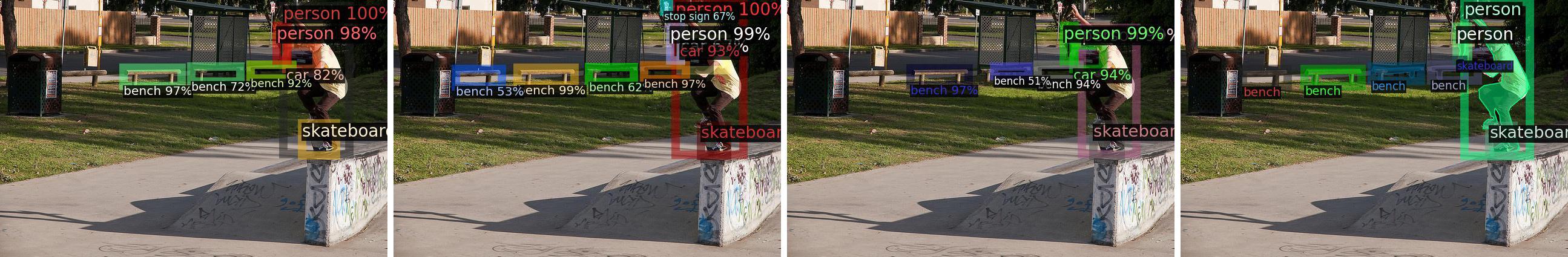}
    \end{subfigure}\\
    \begin{subfigure}[t]{\linewidth}
         \centering
         \includegraphics[width=\linewidth]{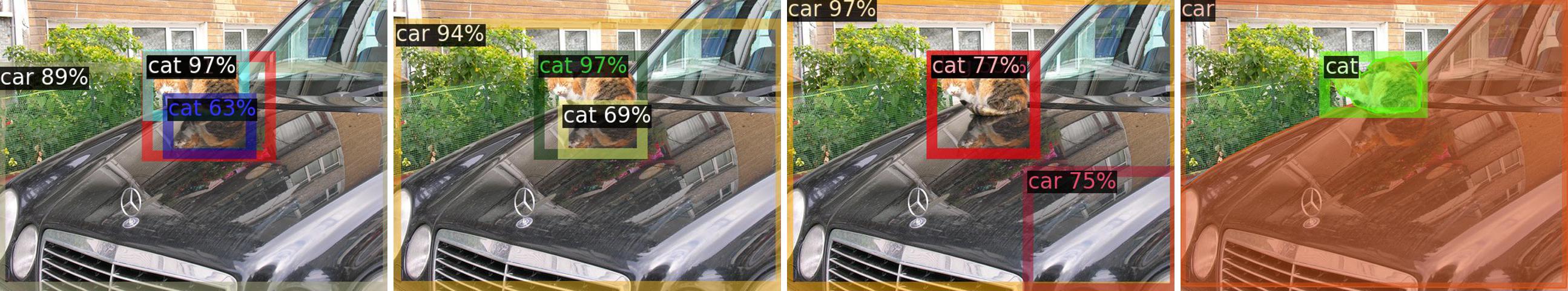}
    \end{subfigure}
    \caption{From left to right: visualization of the detection results by HTC, `HTC + RFP', `HTC + SAC' and the ground truth.
    }
    \label{fig:vis}
\end{figure*}

\section{Experiments}\label{sec:exp}

\subsection{Experimental Details}
We conduct experiments on COCO dataset~\cite{coco}.
All the models presented in the paper are trained on the split of \texttt{train2017} which has 115k labeled images.
Then, we test the models on \texttt{val2017} and \texttt{test-dev}.
We implement DetectoRS with mmdetection~\cite{mmdetection}.
Our baseline model is HTC~\cite{htc}, which uses the bounding box and instance segmentation annotations from the dataset.
Runtime is measured on a single NVIDIA TITAN RTX graphics card.
We strictly follow the experimental settings of HTC~\cite{htc}.
For ablation studies, we train models for 12 epochs with the learning rate multiplied by 0.1 after 8 and 12 epochs.
Additionally, other training and testing settings are kept the same and no bells and whistles are used for them.
For our main results after the ablation studies, we use multi-scale training with the long edge set to 1333 and the short edge randomly sampled from [400, 1200].
We train the models for 40 epochs with the learning rate multiplied by 0.1 after 36 and 39 epochs.
Soft-NMS~\cite{softnms} is used for ResNeXt-101-32x4d and ResNeXt-101-64x4d.
We also report the results with and without test-time augmentation (TTA), which includes horizontal flip and multi-scale testing with the short edge set to [800, 1000, 1200, 1400, 1600] and the long edge set to 1.5x short edge.

\subsection{Ablation Studies}

\begin{table}[t]
\small
\setlength{\tabcolsep}{0.32em}

    \centering
    \begin{tabular}{l|cccccc}
    \toprule
    & AP & AP$_{50}$ & AP$_{75}$ & AP$_{S}$ & AP$_{M}$ & AP$_{L}$ \\
    \midrule
    Baseline HTC & 42.0 & 60.8 & 45.5 & 23.7 & 45.5 & 56.4 \\
    \midrule
    RFP & 46.2 & 65.1 & 50.2 & 27.9 & 50.3 & 60.3 \\
    RFP + sharing & 45.4 & 64.1 & 49.4 & 26.5 & 49.0 & 60.0 \\
    RFP - aspp & 45.7 & 64.2 & 49.6 & 26.7 & 49.3 & 60.5 \\
    RFP - fusion & 45.9 & 64.7 & 50.0 & 27.0 & 50.1 & 60.1 \\
    RFP + 3X & 47.5 & 66.3 & 51.8 & 29.0 & 51.6 & 61.9 \\
    \midrule
    SAC & 46.3 & 65.8 & 50.2 & 27.8 & 50.6 & 62.4 \\
    SAC - DCN & 45.3 & 65.0 & 49.3 & 27.5 & 48.7 & 60.6 \\
    SAC - DCN - global & 44.3 & 63.7 & 48.2 & 25.7 & 48.0 & 59.6 \\
    SAC - DCN - locking & 44.7 & 64.4 & 48.7 & 26.0 & 48.7 & 59.0 \\
    SAC - DCN + DS & 45.1 & 64.6 & 49.0 & 26.3 & 49.3 & 60.1 \\
    \bottomrule
    \end{tabular}
    \caption{Ablation study of RFP (the middle group) and SAC (the bottom group) on COCO \texttt{val2017} with ResNet-50.}
    \label{tab:abl}   

\end{table}

In this subsection, we show the ablation studies of RFP and SAC in \tabref{tab:rfp_sac} and \tabref{tab:abl}.
\tabref{tab:rfp_sac} shows the box and mask AP of the baseline HTC with ResNet-50 and FPN as its backbone.
Then, we add our proposed RFP and SAC to the baseline HTC, both of which are able to improve AP by $>4$\% without too much decrease in the speed.
Combining them together results in our DetectoRS which achieves $49$\% box AP and 42.1\% mask AP at 3.9 fps.

\tabref{tab:abl} shows the individual ablation study of RFP and SAC where we present the sources of their improvements.
For RFP,
we show `RFP + sharing' where $B_i^{1}$ and $B_i^{2}$ share their weights.
We also demonstrate the improvements of the ASPP module and the fusion module by presenting the performance of RFP without them as in `RFP - aspp' and `RFP - fusion'.
Finally, we increase the unrolled step $T$ from 2 to 3 and get `RFP + 3X', which further improves the box AP by $1.3$\%.
For SAC, we first experiment with SAC without DCN~\cite{dcn} (\ie, `SAC - DCN').
Then, we show that the global context is able to bring improvements on AP in `SAC - DCN - global'.
`SAC - DCN - locking' breaks the locking mechanism in \figref{fig:sac} where the second convolution uses only $\Delta\mathbf{w}$, proving that weight locking is necessary for SAC.
Finally, in `SAC - DCN + DS (dual-switch)', we replace $\mathbf{S}(x)$ and $1-\mathbf{S}(x)$ with two independent switches $\mathbf{S}_1(x)$ and $\mathbf{S}_2(x)$.
The ablation study in \tabref{tab:abl} shows that the formulations of RFP and SAC have the best configuration within the design space we have explored.

\begin{table*}
\small
\renewcommand{\arraystretch}{0.9}
\setlength{\tabcolsep}{0.8em}
    \centering
    \begin{tabular}{l|cc|ccc|ccc}
    \toprule
    Method & Backbone & TTA & AP$_{\text{bbox}}$ & AP$_{50}$ & AP$_{75}$ & AP$_{S}$ & AP$_{M}$ & AP$_{L}$ \\
    \midrule
    YOLOv3~\cite{yolov3} & DarkNet-53 & & 33.0 & 57.9 & 34.4 & 18.3 & 25.4 & 41.9 \\
    RetinaNet~\cite{retinanet} & ResNeXt-101 & & 40.8 & 61.1 & 44.1 & 24.1 & 44.2 & 51.2 \\
    RefineDet~\cite{refinedet} & ResNet-101 & \checkmark & 41.8 & 62.9 & 45.7 & 25.6 &  45.1 & 54.1 \\
    CornerNet~\cite{cornernet} & Hourglass-104 & \checkmark & 42.1 &  57.8 & 45.3 & 20.8 & 44.8 & 56.7 \\
    ExtremeNet~\cite{extremenet} & Hourglass-104 & \checkmark & 43.7 & 60.5 & 47.0 & 24.1 & 46.9 & 57.6 \\
    FSAF~\cite{fsaf} & ResNeXt-101 & \checkmark & 44.6 & 65.2 & 48.6 & 29.7 & 47.1 & 54.6 \\
    FCOS~\cite{tian2019fcos} & ResNeXt-101 & & 44.7 & 64.1 & 48.4 & 27.6 & 47.5 & 55.6 \\
    CenterNet~\cite{centernet} & Hourglass-104 & \checkmark & 45.1 & 63.9 &  49.3 & 26.6 & 47.1 & 57.7 \\
    NAS-FPN~\cite{nasfpn} & AmoebaNet & & 48.3 & - & - & - & - & - \\
    SEPC~\cite{sepc} & ResNeXt-101 & & 50.1 & 69.8 & 54.3 & 31.3 & 53.3 & 63.7 \\
    SpineNet~\cite{du2019spinenet} & SpineNet-190 & & 52.1 & 71.8 & 56.5 & 35.4 & 55.0 & 63.6 \\
    EfficientDet-D7~\cite{efficientdet} & EfficientNet-B6 & & 52.2 & 71.4 & 56.3  & - & - & - \\
    EfficientDet-D7x (Model Zoo on GitHub) & - & - & 55.1 & 74.3 & 59.9 & 37.2 & 57.9 & 68.0 \\
    \midrule
    Mask R-CNN~\cite{maskrcnn} & ResNet-101 & & 39.8 & 62.3 & 43.4 & 22.1 & 43.2 & 51.2 \\
    Cascade R-CNN~\cite{cascadercnn} & ResNet-101 & & 42.8 & 62.1 & 46.3 & 23.7 & 45.5 & 55.2 \\
    Libra R-CNN~\cite{pang2019libra} & ResNeXt-101 & & 43.0 & 64.0 & 47.0 & 25.3 & 45.6 & 54.6 \\
    DCN-v2~\cite{dcnv2} & ResNet-101 & \checkmark & 46.0 & 67.9 & 50.8 & 27.8 & 49.1 & 59.5 \\
    PANet~\cite{panet} & ResNeXt-101 & & 47.4 & 67.2 & 51.8 & 30.1 & 51.7 & 60.0 \\
    SINPER~\cite{singh2018sniper} & ResNet-101 & \checkmark & 47.6 & 68.5 & 53.4 & 30.9 & 50.6 & 60.7 \\
    SNIP~\cite{singh2018analysis} & Model Ensemble & \checkmark & 48.3 & 69.7 & 53.7 & 31.4 & 51.6 & 60.7 \\
    TridentNet~\cite{tridentnet} &  ResNet-101 & \checkmark & 48.4 & 69.7 & 53.5 & 31.8 & 51.3 & 60.3 \\ 
    Cascade Mask R-CNN~\cite{cascadercnn} & ResNeXt-152 & \checkmark & 50.2 & 68.2 & 54.9 & 31.9 & 52.9 & 63.5 \\
    TSD~\cite{tsd} & SENet154 & \checkmark & 51.2 & 71.9 & 56.0 & 33.8 & 54.8 & 64.2 \\
    MegDet~\cite{peng2018megdet} & Model Ensemble & \checkmark & 52.5 & - & - & - & - & - \\
    CBNet~\cite{liu2019cbnet} & ResNeXt-152 & \checkmark & 53.3 & 71.9 & 58.5 & 35.5 & 55.8 & 66.7 \\
    \midrule
    HTC~\cite{htc} & ResNet-50 & & 43.6 & 62.6 & 47.4 & 24.8 & 46.0 & 55.9 \\
    HTC & ResNeXt-101-32x4d & & 46.4 & 65.8 & 50.5 & 26.8 & 49.4 & 59.6 \\
    HTC & ResNeXt-101-64x4d & & 47.2 & 66.5 & 51.4 & 27.7 & 50.1 & 60.3 \\
    HTC + DCN~\cite{dcn} + multi-scale training & ResNeXt-101-64x4d &  & 50.8 & 70.3 & 55.2 & 31.1 & 54.1 & 64.8 \\
    \midrule
    DetectoRS & ResNet-50 & & 51.3 & 70.1 & 55.8 & 31.7 & 54.6 & 64.8 \\
    DetectoRS & ResNet-50 & \checkmark & 53.0 & 72.2 & 57.8 & 35.9 & 55.6 & 64.6 \\
    DetectoRS & ResNeXt-101-32x4d &  & 53.3 & 71.6 & 58.5 & 33.9 & 56.5 & 66.9 \\
    DetectoRS & ResNeXt-101-32x4d & \checkmark & 54.7 & 73.5 & 60.1 & 37.4 & 57.3 & 66.4 \\
    DetectoRS & ResNeXt-101-64x4d & \checkmark & 55.7 & 74.2 & 61.1 & 37.7 & 58.4 & 68.1 \\ %
    \bottomrule
    \end{tabular}
    \caption{State-of-the-art comparison on COCO \texttt{test-dev} for bounding box object detection. TTA: test-time augmentation, which includes multi-scale testing, horizontal flipping, \etc.
    The input size of DetectoRS without TTA is (1333, 800).}
    \label{tab:bbox}
\end{table*}

\begin{figure}
  \includegraphics[width=\linewidth]{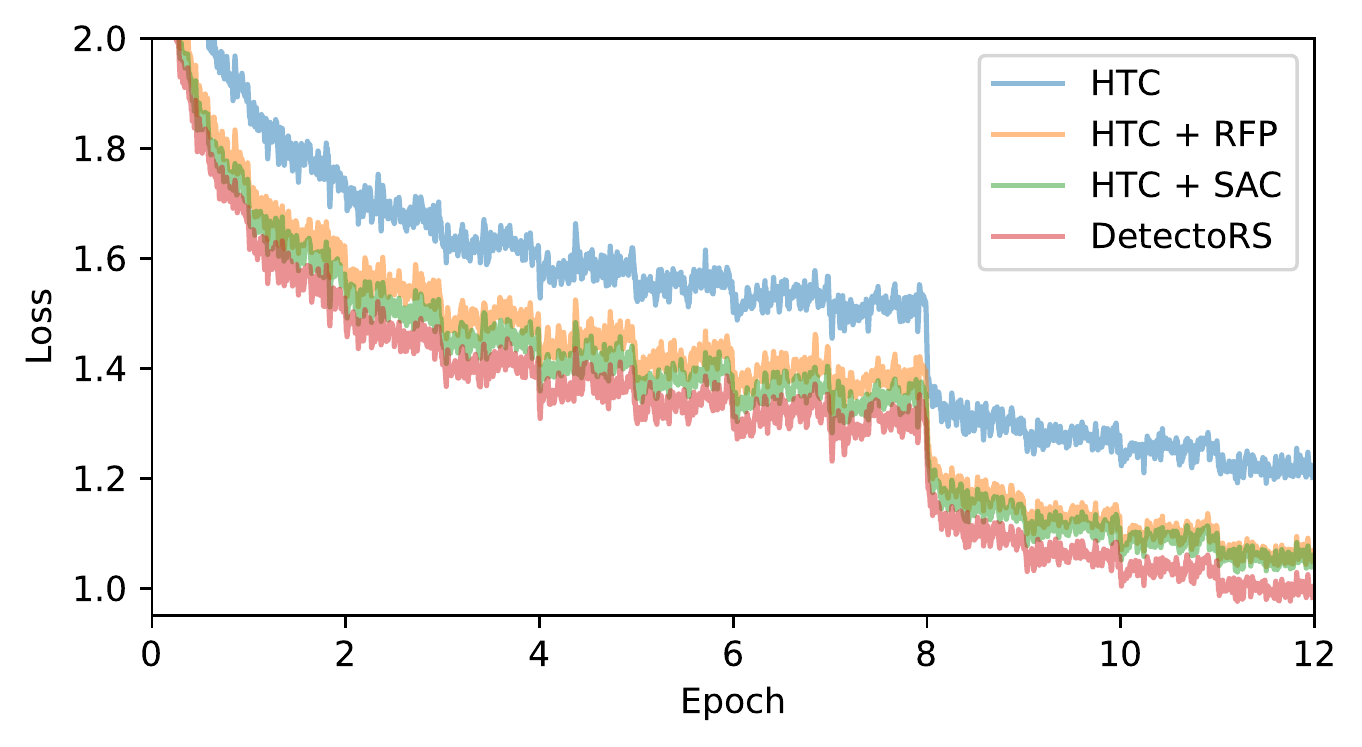}
  \caption{Comparing training losses of HTC, `HTC + RFP', `HTC + SAC', and DetectoRS during 12 training epochs.}
  \label{fig:loss}%
\end{figure}

\figref{fig:vis} provides visualization of the results by HTC, `HTC + RFP' and `HTC + SAC'.
From this comparison,
we notice that RFP, similar to human visual perception that selectively enhances or suppresses neuron activations, is able to find occluded objects more easily for which the nearby context information is more critical.
SAC, because of its ability to increase the field-of-view as needed, is more capable of detecting large objects in the images.
This is also consistent with the results of SAC shown in \tabref{tab:rfp_sac} where it has a higher AP$_L$.
\figref{fig:loss} shows the training losses of HTC, `HTC + RFP', `HTC + SAC', and DetectoRS.
Both are able to significantly accelerate the training process and converge to lower losses.

\subsection{Main Results}

In this subsection, we show the main results of DetectoRS.
We equip the state-of-art detector HTC with DetectoRS, and use ResNet-50 and ResNeXt-101 as the backbones for DetectoRS.
The bounding box detection results are shown in \tabref{tab:bbox}.
The results are divided into 4 groups.
The first group shows one-stage detectors.
The second group shows multi-stage detectors.
The third group is HTC, which is the baseline of DetectoRS.
The fourth group is our results.
The results can be also categorized as simple test results and TTA results, where TTA is short for test-time augmentation.
The third column shows whether TTA is used.
Note that different methods use different TTA strategies.
For example, CBNet uses a strong TTA strategy, which can improve their box AP from 50.7\% to 53.3\%.
Our TTA strategy only brings 1.4\% improvement when using ResNeXt-101-32x4d as backbone.
The simple test settings can also vary significantly among different detectors.
DetectoRS uses (1333, 800) as the test image size.
Larger input sizes tend to bring improvements (see \cite{efficientdet}).
DetectoRS adopts the same setting of HTC.

\begin{figure*}
    \centering
    \begin{subfigure}[t]{0.24\linewidth}
        \centering
        \includegraphics[width=\linewidth]{figures/switch/000000007108.jpg}
    \end{subfigure}
    \hfill
    \begin{subfigure}[t]{0.244\linewidth}
        \centering
        \includegraphics[width=\linewidth]{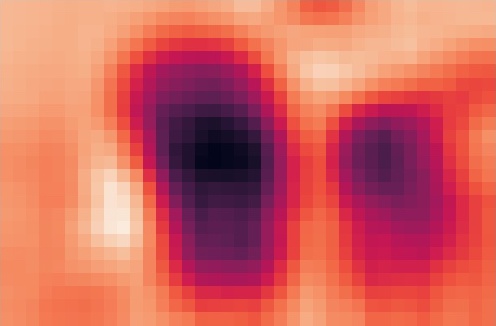}
    \end{subfigure}
    \hfill
        \begin{subfigure}[t]{0.24\linewidth}
        \centering
        \includegraphics[width=\linewidth]{figures/switch/000000017178.jpg}
    \end{subfigure}
    \hfill
    \begin{subfigure}[t]{0.244\linewidth}
        \centering
        \includegraphics[width=\linewidth]{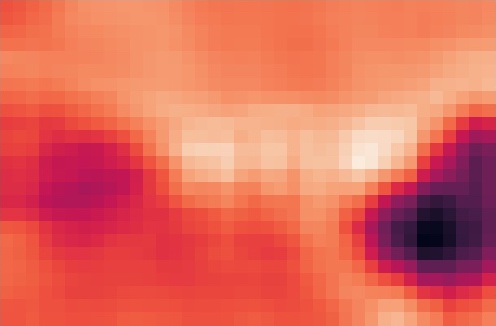}
    \end{subfigure}\\
    \begin{subfigure}[t]{0.24\linewidth}
        \centering
        \includegraphics[width=\linewidth]{figures/switch/000000018491.jpg}
    \end{subfigure}
    \hfill
    \begin{subfigure}[t]{0.244\linewidth}
        \centering
        \includegraphics[width=\linewidth]{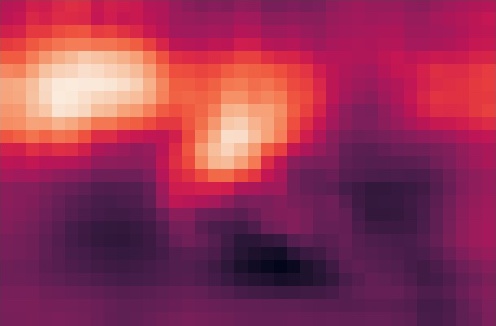}
    \end{subfigure}
    \hfill
    \begin{subfigure}[t]{0.24\linewidth}
        \centering
        \includegraphics[width=\linewidth]{figures/switch/000000018737.jpg}
    \end{subfigure}    
    \hfill
    \begin{subfigure}[t]{0.244\linewidth}
        \centering
        \includegraphics[width=\linewidth]{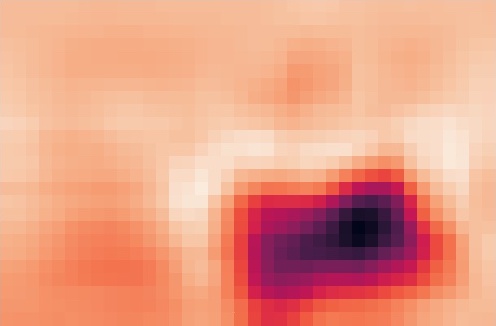}
    \end{subfigure}
    \caption{Visualizing the outputs of the learned switch functions in Switchable Atrous Convolution.
    Darker intensity means that the switch function for that region gathers more outputs from the larger atrous rate.}
    \label{fig:switch}
\end{figure*}

\begin{table*}[]
\small
\renewcommand{\arraystretch}{0.9}
\setlength{\tabcolsep}{0.54em}
    \centering
    \begin{tabular}{l|cc|ccc|ccc}
    \toprule
    Method & Backbone & TTA & AP$_{\text{mask}}$ & AP$_{50}$ & AP$_{75}$ & AP$_{S}$ & AP$_{M}$ & AP$_{L}$ \\
    \midrule
    HTC~\cite{htc} & ResNet-50 & & 38.5 & 60.1 & 41.7 & 20.4 & 40.6 & 51.2 \\
    HTC & ResNeXt-101-32x4d & & 40.7 & 63.2 & 44.1 & 22.0 & 43.3 & 54.2 \\
    HTC & ResNeXt-101-64x4d & & 41.3 & 63.9 & 44.8 & 22.7 & 44.0 & 54.7 \\
    HTC + DCN~\cite{dcn} +  multi-scale training & ResNeXt-101-64x4d &  & 44.2 & 67.8 & 48.1 & 25.3 & 47.2 & 58.7 \\
    \midrule
    DetectoRS & ResNet-50 & & 44.4 & 67.7 & 48.3 & 25.6 & 47.5 & 58.3 \\
    DetectoRS & ResNet-50 & \checkmark & 45.8 & 69.8 & 50.1 & 29.2 & 48.3 & 58.2 \\
    DetectoRS & ResNeXt-101-32x4d &  & 45.8 & 69.2 & 50.1 & 27.4 & 48.7 & 59.6 \\
    DetectoRS & ResNeXt-101-32x4d & \checkmark & 47.1 & 71.1 & 51.6 & 30.3 & 49.5 & 59.6 \\
    DetectoRS & ResNeXt-101-64x4d & \checkmark & 48.5 & 72.0 & 53.3 & 31.6 & 50.9 & 61.5 \\%
    \bottomrule
    \end{tabular}
    \caption{Instance segmentation comparison on COCO \texttt{test-dev}.}
    \label{tab:segm}
\end{table*}

\begin{table}
 \small
 \renewcommand{\arraystretch}{0.9}
 \setlength{\tabcolsep}{0.8em}
    \centering
    \begin{tabular}{l|c|ccc}
    \toprule
    Method & TTA & PQ & PQ$^{\text{Th}}$ & PQ$^{\text{St}}$ \\
    \midrule
    DeeperLab~\cite{yang2019deeperlab} & & 34.3 & 37.5 & 29.6 \\
    SSAP~\cite{gao2019ssap} & \checkmark & 36.9 & 40.1 & 32.0 \\
    Panoptic-DeepLab~\cite{panopticdeeplab} & \checkmark & 41.4 & 45.1 & 35.9 \\
    Axial-DeepLab-L~\cite{axialdeeplab} & \checkmark & 44.2 & 49.2 & 36.8 \\
    \midrule
    TASCNet~\cite{li2018learning} & & 40.7 & 47.0 & 31.0 \\
    Panoptic-FPN~\cite{panopticfpn} & & 40.9 & 48.3 & 29.7 \\
    AdaptIS~\cite{sofiiuk2019adaptis} & \checkmark & 42.8 & 53.2 & 36.7 \\
    AUNet~\cite{aunet} & & 46.5 & 55.8 & 32.5 \\
    UPSNet~\cite{xiong2019upsnet} & \checkmark & 46.6 & 53.2 & 36.7 \\
    Li \textit{et al.}~\cite{li2020unifying} & & 47.2 & 53.5 & 37.7 \\
    SpatialFlow~\cite{chen2019spatialflow} & \checkmark & 47.3 & 53.5 & 37.9 \\
    SOGNet~\cite{yang2019sognet} & \checkmark & 47.8 & - & - \\
    \midrule
    DetectoRS & \checkmark & 50.0 & 58.5 & 37.2 \\ %
    \bottomrule
    \end{tabular}
  \caption{State-of-the-art comparison on COCO \texttt{test-dev} for panoptic segmentation.}
  \label{tab:pano}
\end{table}

We also show the instance segmentation results in \tabref{tab:segm}.
As many methods in \tabref{tab:bbox} do not provide mask AP in their paper, we only compare DetectoRS with its baseline HTC.
The experimental settings for bounding box and mask object detection are the same except that we report AP$_{\text{mask}}$ instead of AP$_{\text{bbox}}$.
From \tabref{tab:segm}, we can see that consistent with the bounding box results, DetectoRS also brings significant improvements over its baseline for instance segmentation.

Finally, the panoptic segmentation results are presented in \tabref{tab:pano}.
As DetectoRS only detects things, we use the stuff predictions by DeepLabv3+~\cite{deeplabv3plus2018} with backbone Wide-ResNet-41~\cite{wrn2016wide,wu2019wider,sslpanopticdeeplab2020}.
Combing the thing and the stuff predictions using the script available in panoptic API~\cite{kirillov2019panoptic} without tuning any hyper-parameters, we set a new state-of-the-art of $50.0$\% PQ for panoptic segmentation on COCO.

\subsection{Visualizing Learned Switches}
\figref{fig:switch} shows the visualization results of the outputs of the last switch function of `SAC - DCN' in \tabref{tab:abl}.
Darker intensity in the figure means that the switch function for that region gathers more outputs from the larger atrous rate.
Comparing the switch outputs with the original images, we observe that the switch outputs are well aligned with the ground-truth object scales.
These results prove that the behaviours of Switchable Atrous Convolution are consistent with our intuition, which tend to use larger atrous rates when encountering large objects.

\section{Conclusion}

In this paper, motivated by the design philosophy of looking and thinking twice, we have proposed DetectoRS, which includes Recursive Feature Pyramid and Switchable Atrous Convolution.
Recursive Feature Pyramid implements thinking twice at the macro level, where the outputs of FPN are brought back to each stage of the bottom-up backbone through feedback connections.
Switchable Atrous Convolution instantiates looking twice at the micro level, where the inputs are convolved with two different atrous rates.
DetectoRS is tested on COCO for object detection, instance segmentation and panoptic segmentation.
It sets new state-of-the-art results on all these tasks.

{
\small
\subsection*{Acknowledgements}
\noindent The animal and lock icons are made by
\href{http://www.freepik.com/}{Freepik} from \href{https://www.flaticon.com/}{Flaticon}.
The switch icon is made by \href{https://www.flaticon.com/authors/pixel-perfect}{Pixel perfect} from \href{https://www.flaticon.com/}{Flaticon}.
The express icon is made by \href{https://www.flaticon.com/free-icon/conveyor_2897537}{Nhor Phai} from \href{https://www.flaticon.com/}{Flaticon}.
}

{\small
\bibliographystyle{ieee_fullname}
\bibliography{det}
}

\end{document}